\documentclass[submission,copyright,creativecommons]{eptcs}
\providecommand{\event}{SLPCS 2016} 

\title{Coordination in Categorical Compositional\\Distributional Semantics}
\author{Dimitri Kartsaklis
\institute{Queen Mary University of London\\
School of Electronic Engineering and Computer Science\\
Mile End Road, London E1 4NS, United Kingdom
}
\email{d.kartsaklis@qmul.ac.uk} }

\def\titlerunning{Coordination in Categorical Compositional Distributional Semantics}
\def\authorrunning{D. Kartsaklis}

\usepackage{latexsym}
\usepackage{lipsum}

\usepackage{amsmath}
\usepackage{amsfonts}
\usepackage{amssymb} 
\usepackage{amsthm}
\usepackage{enumerate}

\usepackage[svgnames]{xcolor}
\usepackage{tikz}

\usepackage{gb4e}

\pgfdeclarelayer{edgelayer}
\pgfdeclarelayer{nodelayer}
\pgfsetlayers{edgelayer,nodelayer,main}

\tikzstyle{none}=[inner sep=0pt]
\tikzstyle{plain}=[inner sep=0pt]
\tikzstyle{every picture}=[baseline=(current bounding box).east,scale=0.35,node distance=5mm]

\newcommand{\ov}{\overrightarrow} 
\newcommand{\ten}{\otimes}
\newcommand{\ol}{\overline}
\newcommand{\B}{\mathbf}
\newcommand{\M}{\mathcal}
\DeclareMathOperator{\conj}{CONJ}

\begin{document}
\maketitle

\begin{abstract}
An open problem with categorical compositional distributional semantics is the representation of words that are considered semantically vacuous from a distributional perspective, such as determiners, prepositions, relative pronouns or coordinators. This paper deals with the topic of coordination between identical syntactic types, which accounts for the majority of coordination cases in language. By exploiting the compact closed structure of the underlying category and Frobenius operators canonically induced over the fixed basis of finite-dimensional vector spaces, we provide a morphism as representation of a coordinator tensor, and we show how it lifts from atomic types to compound types. Linguistic intuitions are provided, and the importance of the Frobenius operators as an addition to the compact closed setting with regard to language is discussed.
\end{abstract}

\section{Introduction}

Inspired by quantum mechanics, and specifically by the category-theoretic manifestation of quantum mechanics as set out by Abramsky and Coecke \cite{abramsky2004}, the compositional framework of Coecke, Sadrzadeh and Clark \cite{Coeckeetal} provides an intuitive way to model the interactions between words within a sentence in the context of a distributional model of meaning. In this setting, nouns are represented as simple vectors living in a basic vector space, while words with a relational nature, such as verbs and adjectives, are multi-linear maps living in tensor product spaces and acting on the noun vectors. The model uses the framework of compact closed categories to unify a grammar expressed as a Lambek pregroup \cite{Lambek} with the category of finite-dimensional vector spaces and linear maps ($\B{FdVect}$), which is where a distributional model of semantics lives; this is achieved by means of a functorial passage that essentially translates every grammatical derivation to a multi-linear algebraic manipulation on the word vectors and tensors. 

A long-standing challenge in compositional distributional models is the representation of functional words, such as prepositions, relative pronouns, or coordinators, since in principle the meaning of these cannot be given by distributional or any other statistical methods. In order to cope with this problem, researchers in the past have exploited the notion of Frobenius algebras which can be canonically induced over the basis of finite-dimensional vector spaces \cite{coecke2006}. Frobenius algebras constitute one of the fundamental structures of categorical quantum mechanics \cite{abramsky2004} (the other one being compact closed categories), where they are used to model classical operations, such as copying or deleting information, that are not allowed in the quantum world. Interestingly, these properties of Frobenius algebras have also been proved useful in language, modelling linguistic aspects for which the representation power of the framework of compact closed categories is insufficient \cite{relpronouns1,relpronouns2,kartsaklis2012,kartsaklis-mol,coecke2016}. 

In this paper we show how Frobenius algebras, in conjunction with the compact closed structure of $\B{FdVect}$, can be used to model one of the most ubiquitous phenomena in language, that of coordination.\footnote{A preliminary account of this subject can be found in the doctoral thesis of the author \cite{kartsaklis-dphil}.} As we will see in Section \ref{sec:frob-intuition}, from a linguistic perspective Frobenius multiplication can be seen as enforcing the two inputs to contribute equally to the result---an idea that captures the essence of coordination. Furthermore, the Frobenius co-multiplication allows the duplication of information when this makes linguistic sense; as it will become evident in Section \ref{sec:coord-compound}, this action is necessary when coordinating text constituents with compound grammatical types, such as verb phrases. The coordinating morphisms we provide in this paper make use of these two notions; furthermore, they get intuitive linear-algebraic interpretations and come with compact composition formulas that simplify their  practical application. Finally, in Section \ref{sec:stripping} we briefly examine the application of the Frobenius machinery on non-standard coordination cases that involve certain forms of ellipsis. 

\section{Background}

This section provides a very short introduction to the compositional framework of Coecke et al. \cite{Coeckeetal} and Frobenius algebras over $\B{FdVect}$ \cite{coecke2006}; the interested reader is encouraged to refer to the original papers for more details. We furthermore assume familiarity with the graphical language of compact closed categories; for an introduction specific to linguistics, see \cite{piedeleu2015open-calco}, App. A.

\subsection{A functorial passage from syntax to semantics}

We recall that a pregroup algebra is a partially ordered monoid, each element $p$ of which has a left and a right adjoint, denoted as $p^l$ and $p^r$ respectively, such that:

\vspace{-0.2cm}
\begin{equation}
  p\cdot p^r \leq 1 \leq p^r \cdot p \quad\quad\quad\quad p^l\cdot p \leq 1 \leq p\cdot p^l
  \label{equ:pregroups}
\end{equation}

Take $P(\M{B})$ to be a pregroup algebra generated over a set of basic (or {\em atomic}) grammatical types $\M{B}$, and $\Sigma$ the vocabulary of the language; then a pregroup grammar is a relation $\Sigma \times P(\M{B})$, denoted as $P(\Sigma,\M{B})$ that assigns a grammar type to every word in the vocabulary \cite{Lambek}. Assuming $\M{B}=\{s,n\}$, where $s$ stands for a well-formed sentence and $n$ for a well-formed noun phrase, we say that a word sequence $w_1w_2\hdots w_n$ forms a grammatical sentence whenever $t_1\cdot t_2 \cdot \hdots \cdot t_n \leq s$, for $(w_i,t_i)\in P(\Sigma,\M{B})$. For example, given the type assignments (`Mary', $n$), (`likes', $n^r\cdot s \cdot n^l$), (`musicals', $n$), the sentence ``Mary likes musicals'' is grammatical since $n\cdot n^r\cdot s \cdot n^l\cdot n \leq 1\cdot s \cdot 1 \leq s$, according to Equation \ref{equ:pregroups}. Note that the type of the transitive verb `likes' is {\em compound}, denoting a word that expects noun phrases at both of its sides in order to return a sentence. 

Both a pregroup grammar and $\B{FdVect}$ have compact closed structure. For a pregroup, the structural morphisms of compact closure $\epsilon$ and $\eta$ become:

\vspace{-0.2cm}
\begin{equation}
   \epsilon^r : p\cdot p^r \leq 1~~,~~\epsilon^l: p^l\cdot p \leq 1 
   ~~~~~~~~~~~~~~~~~
   \eta^r: 1\leq p^r \cdot p~~,~~\eta^l: 1\leq p\cdot p^l
   \label{equ:pregroup-maps}
\end{equation}

Unlike a pregroup, $\B{FdVect}$ is a symmetric compact closed category, meaning that for every pair of objects $A,B$ there is an isomorphism $\sigma: A\ten B \cong B\ten A$. In $\B{FdVect}$, the $\epsilon$ map becomes the inner product between the involved vectors, and  the $\eta$ map defines identity matrices:

\vspace{-0.2cm}
\begin{equation}
\label{equ:innerprod}
  \epsilon^l = \epsilon^r: V \otimes V \to \mathbb{R}:: \sum\limits_{ij} c_{ij}(\ov{v_i} \otimes \ov{v_j}) \mapsto \sum\limits_{ij} c_{ij}\langle \ov{v_i}|\ov{v_j}\rangle ~~~~~
   \eta^l = \eta^r: \mathbb{R} \to V \otimes V :: 1 \mapsto \sum\limits_i \ov{v_i} \otimes \ov{v_i}
\end{equation}

A structure-preserving passage from syntax to semantics can be developed between these two categories by using a strongly monoidal functor:

\begin{equation}
    \M{F}: P(\M{B},\Sigma) \to \B{FdVect}
\end{equation}

\noindent that sends each atomic type $x$ to a vector space $X$, and compound types to tensor product of spaces, since $\M{F}(p\cdot q) = \M{F}(p)\ten \M{F}(q)$. 

As an example, consider the sentence ``John sleeps''. The pregroup derivation takes the form $n\cdot n^r\cdot s \leq 1\cdot s \leq s$, translated to morphism $(\epsilon^r\cdot 1_s)\circ (n\cdot n^r\cdot s)$ according to Equation \ref{equ:pregroup-maps}. Applying our functor to this, and taking $\ov{John} \in N$ and $\ol{sleep} \in N^r\ten S$ to be the semantic representations of the words, will give:

\vspace{-0.3cm}
\begin{eqnarray*}
 \M{F}[(\epsilon^r\cdot 1_s)\circ (n\cdot n^r\cdot s)] & = &
 \M{F}(\epsilon^r\cdot 1_s)\circ \M{F}(n\cdot n^r\cdot s) =
 (\epsilon^r \ten 1_S )\circ (N\ten N^r \ten S) \\ & := &
 (\epsilon^r \ten 1_S )\circ (\ov{John}\ten\ol{sleep}) = \ov{John}^{\mathsf{T}}\times \ol{sleep}
\end{eqnarray*} 

From a linear-algebraic perspective, the above is just the matrix multiplication of the matrix representing the intransitive verb `sleep' with the vector for noun `John'. For higher-order tensors (such as a transitive verb living in $N^r\ten S\ten N^l$), the composition operation generalizes to tensor contraction. 

\subsection{Frobenius algebras}

An object $X$ in a compact closed category ${\cal C}$ has a Frobenius structure on it if there exist morphisms $\Delta \colon X \to X \otimes X$, $\iota \colon X \to I$ and $\mu \colon  X \otimes X \to X$, $\zeta \colon I \to X$ satisfying (among other associativity and unit conditions) the following:

\vspace{-0.3cm}
\begin{equation}
(\mu \otimes 1_X) \circ (1_X \otimes \Delta) \ = \  \Delta \circ \mu  \ = \  (1_X \otimes \mu) \circ (\Delta \otimes 1_X)
\end{equation}

In the category \textbf{FdVect}, any vector space $V$ with a fixed basis $\{\ov{v_i}\}_i$ has a commutative special Frobenius algebra over it, explicitly given as follows \cite{coecke2006}:

\vspace{-0.3cm}
\begin{align}
\label{equ:frob}
  \Delta ::\ov{v_i} \mapsto \ov{v_i} \otimes \ov{v_i} & & \iota::\ov{v_i} \mapsto 1 \nonumber \\ 
 \mu:: \ov{v_i} \otimes \ov{v_j} \mapsto \delta_{ij}\ov{v_i} :=  \left\{\begin{array}{c l}
   \ov{v_i} & i=j \\ 
   \ov{0} & i \neq j
\end{array} \right. & &
  \zeta::1 \mapsto \sum_i \ov{v_i}
\end{align}

For $\ov{u} \in V, \ol{w} \in V \otimes V$, we have that $\Delta(\ov{u})\in V \otimes V$ is a diagonal matrix whose diagonal elements are the weights of $\ov{u}$, and $\mu(\ol{w}) \in V$ is a vector consisting only of the diagonal elements of $\ol{w}$.\footnote{In this paper the notation $\ov{v}$ refers to a vector, while $\ol{w}$ denotes a tensor of order $> 1$.} For the purposes of this paper we refer to $\Delta$ map as the {\em copying} operation, and to $\mu$ map as the {\em merging} operation.

\section{Linguistic uses of the Frobenius operators}
\label{sec:frob-intuition}

The Frobenius operators in $\B{FdVect}$ as given in Equation \ref{equ:frob} adhere to intuitive interpretations that make them important additions to the underlying compact closed setting. In particular, while the standard $\epsilon$-composition has a transformational effect,\footnote{For example, an intransitive verb is a map $N\to S$, faithfully encoded as a matrix living in $N\ten S$, that transforms the input noun into a sentence.} the $\mu$ map can be seen as an alternative form of composition that imposes equal contribution of the operands to the final result; linear-algebraically, while $\epsilon$-composition is tensor contraction requiring one of the interacting words to be of a higher order than the other, $\mu$-composition takes the form of element-wise multiplication between tensors of the same order:

\begin{equation}
  \mu(\ov{x}\ten \ov{y}) = \sum_{ij} \delta_{ij} x_i y_j \ov{v_{i}} = \sum_i x_i y_i \ov{v_i} =  \ov{x} \odot \ov{y}
\end{equation}

\noindent where $\ov{v_i}$ is a basis vector and $\odot$ denotes element-wise vector multiplication. From a linguistic perspective, there are many cases where such an interaction is desirable. Sadrzadeh et al. \cite{relpronouns1,relpronouns2}, for example, work on nouns modified by relative clauses (e.g. ``The man who likes Mary'') and present a construction for the relative pronoun that results in $\mu$-composing  the vector of the noun (`man') with the vector of the modifying verb phrase (`likes Mary'). Similarly, Kartsaklis and Sadrzadeh \cite{kartsaklis-mol} inject intonational information into a sentence vector by merging the vectors of the theme of the sentence (information that is known to both interlocutors in a conversation) and the rheme of the sentence (information that is new to the addressee). Finally, Coecke and Lewis \cite{coecke2016} argue that $\mu$-composition can approximate to a certain extent the meaning of non-compositional compounds, such as `pet fish'. 

Technically, $\mu$-composition has an intersective effect on the elements of the operands. This is more clear when one uses \textbf{Rel} (the category of sets and relations) as the co-domain of our syntax-to-semantics functor $\M{F}$, instead of \textbf{FdVect}. Recall that elements in finite sets can be seen as basis vectors of free modules over the semi-ring of booleans. In this setting, nouns are represented by vectors corresponding to subsets of the universe of discourse, whose $i$-th component is 1 if the $i$-th element is included in the specific set and 0 otherwise. Furthermore, a verb becomes a relation, represented by an adjacency matrix in which the $(i,j)$-th component is 1 if the relation stands for the pair consisting of the $i$-th element of its domain and the $j$-th element of its codomain. It is clear that in these cases element-wise multiplication of the vectors/tensors corresponds exactly to the intersection of the involved sets or relations. The transition from the truth-theoretic setting of \textbf{Rel} to the real-valued \textbf{FdVect}, where a standard distributional model of semantics lives, results in a form of ``quantitative'' intersection between the components of the vectors that has been proved very effective in a number of standard NLP tasks (see, for example, \cite{lapata2010}).

The Frobenius co-multiplication (the ``copying'' $\Delta$ map) has been also proved useful in linguistic applications. Kartsaklis et al. \cite{kartsaklis2012} use it in order to restore the functorial relationship between grammar and verb tensors of order lower than that dictated by their type. In a more conceptual use of this operation, the relative pronoun construction of Sadrzadeh et al. essentially copies the vector of the noun from its original position and allows the information to ``flow'' at the other side of the pronoun and interact with the vector of the verb phrase, providing a means of syntactic movement. The current paper builds on all the above intuitions, and in the following sections these ideas are applied to model coordination in language.

\section{Coordination in CCDS}

Coordination is perhaps the linguistic aspect in which the notions of merging and copying information find their most natural application. Two coordinated phrases or sentences can be seen as contributing equally to the final outcome; we would expect, for example, the vector of the sentence ``John reads and Mary sleeps'' to reflect equally the vectors of the two coordinated sentences, ``John reads'' and ``Mary sleeps''. Furthermore, distributivity conditions suggest that parts of the coordinate structure should be copied and interact with each one of the conjuncts separately. For example, it is the case that:

\begin{exe}
  \ex\label{ex:np-dist} Mary studies philosophy and history $\models$ Mary studies philosophy and Mary studies history 
  \ex\label{ex:vp-dist} Men like sports and play football $\models$ Men like sports and men play football
\end{exe}

\noindent where the symbol $\models$ denotes entailment. Therefore, merging and copying can be seen as the key processes of coordination, and in this section we will show that an effective use of Frobenius operators in conjunction with the underlying compact closed structure of $\B{FdVect}$ allows us to model a variety of coordination cases in categorical compositional distributional semantics.

For the analysis that follows we consider the usual ternary rule $X~\conj~X \to X$, which states that coordination always takes place between conjuncts of the same type and produces a result that again matches that specific type. In pregroup terms, this is achieved by assigning the type $x^r\cdot x\cdot x^l$ to the conjunction (where $x$ can be an atomic or a compound type), which leads to the following generic derivation:

\begin{equation}
 x\cdot (x^r\cdot x\cdot x^l)\cdot x \leq 1\cdot x\cdot 1 = x
 \label{equ:ternary}
\end{equation}

Taking $\ov{x_1},\ov{x_2} \in \M{F}(x)=X$ to be the semantic representations of the two conjuncts and $\ol{conj}_X \in X^r \ten X \ten X^l$ the coordination tensor,\footnote{Although for a vector space $V$ in $\B{FdVect}$ it is always the case that $V\cong V^r \cong V^l$, in this paper the adjoints of vector spaces will be explicitly stated for additional clarity.}
we translate the above pregroup derivation to \textbf{FdVect} as follows:

\begin{equation}
\M{F}\left[ (\epsilon^r_x \cdot 1_x \cdot \epsilon^l_x) \circ (x \cdot x^r\cdot x \cdot x^l \cdot x) \right] = 
(\epsilon^r_X\ten 1_X \ten \epsilon^l_X) \circ (\ov{x_1}\ten\ol{conj}_X\ten\ov{x_2})
\label{equ:trans}
\end{equation}

Our main concern is to find a way to transform the composition function $\epsilon^r_X\ten 1_X \ten \epsilon^l_X$ (which for the moment is solely expressed in terms of $\epsilon$-maps) to a function that applies $\mu$-composition between the two operands, thus enforces equal contribution of the conjuncts in the final result. The connection between the two ways of composing the conjuncts is made explicit in the following diagram:

\begin{equation}
\footnotesize

\begin{tikzpicture}
	\begin{pgfonlayer}{nodelayer}
		\node [style=none] (0) at (-9.5, 3.5) {$X\ten X$};
		\node [style=none] (1) at (5.5, 3.5) {$X\ten X^r \ten X \ten X \ten X^l \ten X$};
		\node [style=none] (2) at (-8, 3.5) {};
		\node [style=none] (3) at (0.5, 3.5) {};
		\node [style=none] (4) at (5.5, -1.5) {};
		\node [style=none] (5) at (5.5, -2.25) {$X\ten X^r \ten X \ten X^l \ten X$};
		\node [style=none] (6) at (5.5, 2.75) {};
		\node [style=none] (7) at (-4, 4.25) {\scriptsize{$1_X\ten\eta^r_X\ten\eta^l_X\ten 1_X$}};
		\node [style=none] (8) at (-8.5, -2.25) {};
		\node [style=none] (9) at (1.5, -2.25) {};
		\node [style=none] (10) at (-9.5, -2.25) {$X$};
		\node [style=none] (11) at (10.25, 1) {\scriptsize{$1_X\ten 1_{X^r}\ten \mu_X \ten 1_{X^l}\ten 1_X$}};
		\node [style=none] (12) at (-9.5, 2.75) {};
		\node [style=none] (13) at (-9.5, -1.5) {};
		\node [style=none] (14) at (-3, -3) {\scriptsize{$\epsilon^r_X\ten 1_X \ten \epsilon^l_X$}};
		\node [style=none] (15) at (-10.5, 0.75) {\scriptsize{$\mu_X$}};
		\node [style=none] (16) at (-11, 3.5) {};
		\node [style=none] (17) at (-15, 3.5) {};
		\node [style=none] (18) at (-15.75, 3.5) {$I$};
		\node [style=none] (19) at (-13, 4.25) {\scriptsize{$\ov{x_1}\ten\ov{x_2}$}};
	\end{pgfonlayer}
	\begin{pgfonlayer}{edgelayer}
		\draw [->] (2.center) to (3.center);
		\draw [->] (6.center) to (4.center);
		\draw [<-] (8.center) to (9.center);
		\draw [->] (12.center) to (13.center);
		\draw [->] (17.center) to (16.center);
	\end{pgfonlayer}
\end{tikzpicture}}

\normalsize
\end{equation}

\noindent from which we derive:

\vspace{-0.5cm}
\small
\begin{eqnarray}
  \mu_X \circ (\ov{x_1}\ten \ov{x_2}) & = & (\epsilon^r_X\ten 1_X \ten \epsilon^l_X) \circ (1_X\ten 1_{X^r} \ten \mu_X \ten 1_{X^l} \ten 1_X) \circ
  (1_X\ten \eta^r_X \ten \eta^l_X \ten 1_X) \circ (\ov{x_1}\ten \ov{x_2}) \nonumber \\
  & = & (\epsilon^r_X\ten 1_X \ten \epsilon^l_X) \circ \left(1_X \ten \left[(1_{X^r}\ten \mu_X \ten 1_{X^l})\circ (\eta^r_X\ten\eta^l_X)\right]\ten 1_X \right) \circ (\ov{x_1}\ten \ov{x_2}) \nonumber \\
  & = & (\epsilon^r_X\ten 1_X \ten \epsilon^l_X) \circ \left(\ov{x_1} \ten \left[(1_{X^r}\ten \mu_X \ten 1_{X^l})\circ (\eta^r_X\ten\eta^l_X)\right]\ten \ov{x_2} \right)
  \label{equ:proof}
\end{eqnarray}
\normalsize

The morphism between the square brackets defines a state in $X^r\ten X\ten X^l$, and corresponds exactly to the semantic representation of the coordinator we need (see also Equation \ref{equ:trans}) in order to translate $\epsilon$-composition to $\mu$-composition. The coordinator morphism is stated explicitly below:

\begin{equation}
  \ol{conj}_X : I \xrightarrow{\eta^r_X \ten \eta^l_X} X^r \ten X \ten X \ten X^l \xrightarrow{1_{X^r}\ten \mu_X \ten 1_{X^l}} X^r \ten X \ten X^l
  \label{equ:coord}
\end{equation} 

In the following sections we will use the diagrammatic calculus of compact closed categories to demonstrate the application of Equation \ref{equ:coord} in a number of coordination examples; the cases of atomic and compound types are treated separately.

\subsection{Coordinating atomic types}
\label{sec:coord-atomic}

We start with the simple case of coordinating conjuncts of atomic types, that is, conjuncts whose semantic representation is a vector (nouns) or it can be reduced to a vector by $\epsilon$-composition (e.g. noun phrases and sentences). In the graphical language of compact closed categories, the composition of a coordination over noun phrases according to Equation \ref{equ:coord} takes this form:

\begin{equation}
\footnotesize

\begin{tikzpicture}
	\begin{pgfonlayer}{nodelayer}
		\node [style=none, text height=1.5 ex, text depth=0.25 ex] (0) at (21.75, 3.75) {and};
		\node [style=none] (1) at (21.75, 3) {};
		\node [style=none, text height=1.5 ex, text depth=0.25 ex] (2) at (31, 3) {apples};
		\node [style=none, text height=1.5 ex, text depth=0.25 ex] (3) at (34, 3) {oranges};
		\node [style=none, text height=1.5 ex, text depth=0.25 ex] (4) at (16.75, 2.5) {apples};
		\node [style=none, text height=1.5 ex, text depth=0.25 ex] (5) at (26.75, 2.5) {oranges};
		\node [style=none] (6) at (31, 2) {};
		\node [style=none] (7) at (34, 2) {};
		\node [style=none] (8) at (16.75, 1.5) {};
		\node [style=none] (9) at (26.75, 1.5) {};
		\node [style=none] (10) at (20.5, 1.25) {};
		\node [style=none] (11) at (21.25, 1.25) {};
		\node [style=none] (12) at (22.25, 1.25) {};
		\node [style=none] (13) at (23, 1.25) {};
		\node [style=none] (14) at (30, 1) {};
		\node [style=none] (15) at (31, 1) {};
		\node [style=none] (16) at (32, 1) {};
		\node [style=none] (17) at (33, 1) {};
		\node [style=none] (18) at (34, 1) {};
		\node [style=none] (19) at (35, 1) {};
		\node [draw, circle, minimum size=0.15 cm, fill=white, style=none] (20) at (21.75, 0.78) {};
		\node [style=none] (21) at (15.75, 0.5) {};
		\node [style=none] (22) at (16.75, 0.5) {};
		\node [style=none] (23) at (17.75, 0.5) {};
		\node [style=none] (24) at (19.25, 0.5) {};
		\node [style=none] (25) at (20.5, 0.5) {};
		\node [style=none] (26) at (23, 0.5) {};
		\node [style=none] (27) at (24.25, 0.5) {};
		\node [style=none] (28) at (25.75, 0.5) {};
		\node [style=none] (29) at (26.75, 0.5) {};
		\node [style=none] (30) at (27.75, 0.5) {};
		\node [style=none] (31) at (31, 0.25) {};
		\node [style=none] (32) at (34, 0.25) {};
		\node [style=none] (33) at (16.75, -0.25) {};
		\node [style=none] (34) at (20.5, -0.25) {};
		\node [style=none] (35) at (21.75, -0.25) {};
		\node [style=none] (36) at (23, -0.25) {};
		\node [style=none] (37) at (26.75, -0.25) {};
		\node [style=none, text height=1.5 ex, text depth=0.25 ex] (38) at (31, -0.25) {\footnotesize{$N$}};
		\node [style=none, text height=1.5 ex, text depth=0.25 ex] (39) at (34, -0.25) {\footnotesize{$N$}};
		\node [style=none, text height=1.5 ex, text depth=0.25 ex] (40) at (29, -0.5) {$\mapsto$};
		\node [style=none, text height=1.5 ex, text depth=0.25 ex] (41) at (16.75, -0.75) {\footnotesize{$N$}};
		\node [style=none, text height=1.5 ex, text depth=0.25 ex] (42) at (20.5, -0.75) {\footnotesize{$N^r$}};
		\node [style=none, text height=1.5 ex, text depth=0.25 ex] (43) at (21.75, -0.75) {\footnotesize{$N$}};
		\node [style=none, text height=1.5 ex, text depth=0.25 ex] (44) at (23, -0.75) {\footnotesize{$N^l$}};
		\node [style=none, text height=1.5 ex, text depth=0.25 ex] (45) at (26.75, -0.75) {\footnotesize{$N$}};
		\node [style=none] (46) at (31, -0.75) {};
		\node [style=none] (47) at (34, -0.75) {};
		\node [style=none] (48) at (16.75, -1.25) {};
		\node [style=none] (49) at (20.5, -1.25) {};
		\node [style=none] (50) at (21.75, -1.25) {};
		\node [style=none] (51) at (23, -1.25) {};
		\node [style=none] (52) at (26.75, -1.25) {};
		\node [draw, circle, minimum size=0.15 cm, fill=white, style=none] (53) at (32.5, -1.65) {};
		\node [style=none] (54) at (21.75, -2.5) {};
		\node [style=none] (55) at (32.5, -2.75) {};
	\end{pgfonlayer}
	\begin{pgfonlayer}{edgelayer}
		\draw [thick] (53.center) to (55.center);
		\draw [thick, bend left=90, looseness=2.25] (10.center) to (11.center);
		\draw [thick, looseness=0.00] (6.center) to (16.center);
		\draw [thick, looseness=0.00] (26.center) to (36.center);
		\draw [thick, looseness=0.00] (14.center) to (16.center);
		\draw [thick, looseness=0.00] (9.center) to (30.center);
		\draw [thick, looseness=0.00] (24.center) to (1.center);
		\draw [thick, looseness=0.00] (25.center) to (34.center);
		\draw [thick, looseness=0.00] (15.center) to (31.center);
		\draw [thick, looseness=0.00] (24.center) to (27.center);
		\draw [thick, bend left=270] (48.center) to (49.center);
		\draw [thick, bend right=90] (46.center) to (47.center);
		\draw [thick, looseness=0.00] (29.center) to (37.center);
		\draw [thick, looseness=0.00] (1.center) to (27.center);
		\draw [thick, looseness=0.00] (22.center) to (33.center);
		\draw [thick] (10.center) to (25.center);
		\draw [thick, looseness=0.00] (17.center) to (19.center);
		\draw [thick, looseness=0.00] (21.center) to (23.center);
		\draw [thick, looseness=0.00] (8.center) to (23.center);
		\draw [thick, bend left=270, looseness=1.50] (11.center) to (12.center);
		\draw [thick, bend left=90, looseness=2.25] (12.center) to (13.center);
		\draw [thick, looseness=0.00] (7.center) to (19.center);
		\draw [thick, looseness=0.00] (28.center) to (9.center);
		\draw [thick, looseness=0.00] (28.center) to (30.center);
		\draw [thick, looseness=0.00] (21.center) to (8.center);
		\draw [thick, looseness=0.00] (14.center) to (6.center);
		\draw [thick, looseness=0.00] (17.center) to (7.center);
		\draw [thick] (13.center) to (26.center);
		\draw [thick, bend left=270] (51.center) to (52.center);
		\draw [thick, looseness=0.00] (20.center) to (35.center);
		\draw [thick] (50.center) to (54.center);
		\draw [thick, looseness=0.00] (18.center) to (32.center);
	\end{pgfonlayer}
\end{tikzpicture}}

\normalsize
\end{equation}

In this notation, a vector $\ov{v} \in V$ is represented as a state of $V$, i.e. a morphism $\ov{v}: I\to V$ (the triangle denotes that the domain is the monoidal unit). Similarly, tensors are states of tensor product spaces; e.g. the coordinator tensor above is a morphism $I\to N^r\ten N \ten N^l$. The cups ($\cup$) denote $\epsilon$ maps, and the caps ($\cap$) $\eta$ maps; straight line segments denote identities. Composing morphisms amounts to connecting outputs to inputs, while the tensor product is juxtaposition. The dot node corresponds to the Frobenius multiplication. 

As noted before, the morphism of Equation \ref{equ:coord} provides an interface between standard $\epsilon$-composition (tensor contraction) and $\mu$-composition (element-wise vector multiplication), equipping the underlying compositional framework with an additional layer of flexibility. For the above example we have:

\vspace{-0.4cm}
\begin{equation}
  (\epsilon^r_N\ten 1_N \ten \epsilon^l_N) \circ (\ov{apples}\ten\ol{conj}_N\ten\ov{oranges}) = 
  \mu(\ov{apples} \ten \ov{oranges}) = \ov{apples} \odot \ov{oranges}
\end{equation}

The type of a sentence coordinator is $s^r\cdot s \cdot s^l$, leading to a situation very similar to that of the noun phrase case:

\begin{equation}
  \footnotesize
  
\InputIfFileExists{./tikz/coord-sent-new.tikz}{}{\input{.//tikz//coord-sent-new.tikz}}

  \normalsize
  \label{equ:generic-sent}
\end{equation}

Linear-algebraically this results in a combination of $\epsilon$-composition that takes place within the context of each sentence, and $\mu$-composition that merges the two sentences into a single one at the position of the coordinator:

\begin{equation}
   (\ov{men}^{\mathsf{T}}\times \ol{watch}\times\ov{football}) \odot (\ov{women}^{\mathsf{T}}\times \ol{knit})
\end{equation}

\subsection{Coordinating compound types}
\label{sec:coord-compound}

The simple cases addressed in Section \ref{sec:coord-atomic} make use of the merging Frobenius operator, but they do not include any examples for which  duplication of information is necessary. This requirement emerges when one moves to coordination over complex types, which is more interesting and involved. In order to understand how does this work, it would be instructive to examine the way in which the morphism of Equation \ref{equ:coord} lifts to complex types.  We will use as an example the case of verb phrase coordination. Recall that the pregroup type of a verb phrase is $n^r\cdot s$; that is, something that expects a noun (a subject) from the left in order to return a sentence. The semantic representation of a verb phrase in $\B{FdVect}$ is a matrix living in $N^r\ten S\cong N\ten S$; note however that what follows can be directly generalized to tensors of higher order. 

We start by expressing the maps on the compound objects in terms of the atomic objects. For an arbitrary compound object $U\ten V$, recall that:

\begin{equation}
\footnotesize

\begin{tikzpicture}
	\begin{pgfonlayer}{nodelayer}
		\node [style=none] (0) at (-12, 4.75) {$I$};
		\node [style=none] (1) at (-7.25, 4.75) {$V^r \ten V$};
		\node [style=none] (2) at (-11.25, 4.75) {};
		\node [style=none] (3) at (-8.75, 4.75) {};
		\node [style=none] (4) at (-7.25, 0.5) {};
		\node [style=none] (5) at (-7.25, -0.25) {$V^r \ten U^r \ten U \ten V=$};
		\node [style=none] (6) at (-7.25, 4) {};
		\node [style=none] (7) at (-7.25, -1.5) {$(U\ten V)^r \ten (U\ten V)$};
		\node [style=none] (8) at (-10, 5.5) {\scriptsize{$\eta^r_V$}};
		\node [style=none] (9) at (-4.75, 2.5) {\scriptsize{$1_{V^r}\ten \eta^r_U \ten 1_V$}};
		\node [style=none] (10) at (-11.25, 4) {};
		\node [style=none] (11) at (-8, 0.5) {};
		\node [style=none] (12) at (-11, 2) {\scriptsize{$\eta^r_{U\ten V}$}};
		\node [style=none] (13) at (7.5, 0.5) {};
		\node [style=none] (14) at (3.5, 4) {};
		\node [style=none] (15) at (7.5, 4.75) {$U \ten U^l$};
		\node [style=none] (16) at (4.75, 5.5) {\scriptsize{$\eta^l_U$}};
		\node [style=none] (17) at (6.75, 0.5) {};
		\node [style=none] (18) at (7.5, 4) {};
		\node [style=none] (19) at (10, 2.5) {\scriptsize{$1_U \ten \eta^l_V \ten 1_{U^l}$}};
		\node [style=none] (20) at (3.5, 4.75) {};
		\node [style=none] (21) at (3.75, 2) {\scriptsize{$\eta^l_{U\ten V}$}};
		\node [style=none] (22) at (6, 4.75) {};
		\node [style=none] (23) at (7.5, -0.25) {$U\ten V \ten V^l \ten U^l=$};
		\node [style=none] (24) at (2.75, 4.75) {$I$};
		\node [style=none] (25) at (7.5, -1.5) {$(U\ten V) \ten (U\ten V)^l$};
	\end{pgfonlayer}
	\begin{pgfonlayer}{edgelayer}
		\draw [->] (2.center) to (3.center);
		\draw [->] (6.center) to (4.center);
		\draw [->] (10.center) to (11.center);
		\draw [->] (20.center) to (22.center);
		\draw [->] (18.center) to (13.center);
		\draw [->] (14.center) to (17.center);
	\end{pgfonlayer}
\end{tikzpicture}}

\normalsize
\end{equation}

\noindent which is graphically depicted as follows:

\begin{equation}
\footnotesize

\begin{tikzpicture}
	\begin{pgfonlayer}{nodelayer}
		\node [style=none, text height=1.5 ex, text depth=0.25 ex] (0) at (-13.25, -1) {\footnotesize{$(U\otimes V)^r$}};
		\node [text depth=0.25 ex, text height=1.5 ex, style=none] (1) at (-9.75, -1) {\footnotesize{$(U\otimes V)$}};
		\node [style=none] (2) at (-13.25, -0.25) {};
		\node [style=none] (3) at (-9.75, -0.25) {};
		\node [style=none] (4) at (-6.25, -1) {\footnotesize{$V^r$}};
		\node [style=none] (5) at (-5, -1) {\footnotesize{$U^r$}};
		\node [style=none] (6) at (-3, -1) {\footnotesize{$U$}};
		\node [style=none] (7) at (-1.75, -1) {\footnotesize{$V$}};
		\node [style=none] (8) at (-5, -0.25) {};
		\node [style=none] (9) at (-3, -0.25) {};
		\node [style=none] (10) at (-6.25, -0.25) {};
		\node [style=none] (11) at (-1.75, -0.25) {};
		\node [style=none] (12) at (-7.75, 0.5) {\footnotesize{$\mapsto$}};
		\node [style=none] (13) at (12.75, -0.25) {};
		\node [style=none] (14) at (9.5, -0.25) {};
		\node [style=none] (15) at (8, 0.5) {\footnotesize{$\mapsto$}};
		\node [style=none] (16) at (12.75, -1) {\footnotesize{$V^l$}};
		\node [style=none] (17) at (10.75, -0.25) {};
		\node [style=none] (18) at (10.75, -1) {\footnotesize{$V$}};
		\node [style=none] (19) at (2.5, -0.25) {};
		\node [style=none] (20) at (14, -1) {\footnotesize{$U^l$}};
		\node [style=none] (21) at (9.5, -1) {\footnotesize{$U$}};
		\node [text depth=0.25 ex, text height=1.5 ex, style=none] (22) at (2.5, -1) {\footnotesize{$(U\otimes V)$}};
		\node [style=none] (23) at (14, -0.25) {};
		\node [style=none] (24) at (6, -0.25) {};
		\node [style=none, text height=1.5 ex, text depth=0.25 ex] (25) at (6, -1) {\footnotesize{$(U\otimes V)^l$}};
	\end{pgfonlayer}
	\begin{pgfonlayer}{edgelayer}
		\draw [ultra thick, ->, bend left=90, looseness=1.75] (2.center) to (3.center);
		\draw [thick, ->, bend left=90, looseness=1.50] (8.center) to (9.center);
		\draw [thick, ->, bend left=90, looseness=1.50] (10.center) to (11.center);
		\draw [<-, ultra thick, bend left=90, looseness=1.75] (19.center) to (24.center);
		\draw [<-, thick, bend left=90, looseness=1.50] (17.center) to (13.center);
		\draw [<-, thick, bend left=90, looseness=1.50] (14.center) to (23.center);
	\end{pgfonlayer}
\end{tikzpicture}}

\normalsize
\label{equ:eta-lift}
\end{equation}

Furthermore, it is also the case that the Frobenius operators coherently lift to compound objects; for the $\mu$ map, this lifting takes the following form:

\begin{equation}
\footnotesize

\begin{tikzpicture}
	\begin{pgfonlayer}{nodelayer}
		\node [style=none] (0) at (-7.25, 3.5) {$(U\ten V)\ten (U\ten V)$};
		\node [style=none] (1) at (5.5, 3.5) {$U\ten U \ten V \ten V$};
		\node [style=none] (2) at (-3.5, 3.5) {};
		\node [style=none] (3) at (2.5, 3.5) {};
		\node [style=none] (4) at (5.5, -0.75) {};
		\node [style=none] (5) at (5.5, -1.5) {$U\ten V$};
		\node [style=none] (6) at (5.5, 2.75) {};
		\node [style=none] (7) at (-0.5, 4.25) {\scriptsize{$1_U\ten \sigma_{U,V} \ten 1_V$}};
		\node [style=none] (8) at (7.5, 1.25) {\scriptsize{$\mu_U\ten \mu_V$}};
		\node [style=none] (9) at (-3.5, 2.75) {};
		\node [style=none] (10) at (4.75, -0.75) {};
		\node [style=none] (11) at (-0.25, 0.5) {\scriptsize{$\mu_{U\ten V}$}};
	\end{pgfonlayer}
	\begin{pgfonlayer}{edgelayer}
		\draw [->] (2.center) to (3.center);
		\draw [->] (6.center) to (4.center);
		\draw [->] (9.center) to (10.center);
	\end{pgfonlayer}
\end{tikzpicture}}

\normalsize
\end{equation}

\noindent
and is graphically shown below:

\begin{equation}
\footnotesize

\begin{tikzpicture}
	\begin{pgfonlayer}{nodelayer}
		\node [style=none, fill=white, minimum size=0.20 cm, circle, draw, thick] (0) at (-4, 0.25) {};
		\node [text depth=0.25 ex, text height=1.5 ex, style=none] (1) at (-5.75, 2.5) {\footnotesize{$(U\otimes V)$}};
		\node [style=none] (2) at (-4, -1.5) {};
		\node [style=none, text height=1.5 ex, text depth=0.25 ex] (3) at (-2.25, 2.5) {\footnotesize{$(U\otimes V)$}};
		\node [style=none] (4) at (-5.75, 1.75) {};
		\node [style=none] (5) at (-2.25, 1.75) {};
		\node [text depth=0.25 ex, text height=1.5 ex, style=none] (6) at (-4, -2.25) {\footnotesize{$(U\otimes V)$}};
		\node [style=none, text height=1.5 ex, text depth=0.25 ex] (7) at (1.25, 2.75) {\footnotesize{$U$}};
		\node [text depth=0.25 ex, text height=1.5 ex, style=none] (8) at (3.25, 2.75) {\footnotesize{$V$}};
		\node [text depth=0.25 ex, text height=1.5 ex, style=none] (9) at (5, 2.75) {\footnotesize{$U$}};
		\node [style=none, text height=1.5 ex, text depth=0.25 ex] (10) at (7, 2.75) {\footnotesize{$V$}};
		\node [style=none] (11) at (1.25, 2) {};
		\node [style=none] (12) at (5, 2) {};
		\node [style=none] (13) at (7, 2) {};
		\node [style=none] (14) at (3.25, 0.75) {};
		\node [style=none] (15) at (5, 0.75) {};
		\node [style=none] (16) at (3.75, 1.25) {};
		\node [style=none] (17) at (4.5, 1.5) {};
		\node [style=none] (18) at (1.25, 0.75) {};
		\node [style=none] (19) at (7, 0.75) {};
		\node [draw, circle, minimum size=0.15 cm, fill=white, style=none] (20) at (2.25, -0.25) {};
		\node [style=none] (21) at (2.25, -1.5) {};
		\node [style=none] (22) at (6, -1.5) {};
		\node [style=none, fill=white, minimum size=0.15 cm, circle, draw] (23) at (6, -0.25) {};
		\node [style=none] (24) at (-0.25, 0.25) {\footnotesize{$\mapsto$}};
		\node [text depth=0.25 ex, text height=1.5 ex, style=none] (25) at (2.25, -2.25) {\footnotesize{$U$}};
		\node [text depth=0.25 ex, text height=1.5 ex, style=none] (26) at (6, -2.25) {\footnotesize{$V$}};
		\node [style=none] (27) at (4, 1.5) {};
		\node [style=none] (28) at (3.25, 2) {};
		\node [style=none] (29) at (4, 1.5) {};
	\end{pgfonlayer}
	\begin{pgfonlayer}{edgelayer}
		\draw [ultra thick, bend right=90, looseness=1.50] (4.center) to (5.center);
		\draw [ultra thick] (0.center) to (2.center);
		\draw [thick, bend left=15, looseness=1.00] (12.center) to (17.center);
		\draw [thick, bend right, looseness=1.00] (16.center) to (14.center);
		\draw [thick] (11.center) to (18.center);
		\draw [thick] (13.center) to (19.center);
		\draw [thick, bend right=90, looseness=1.50] (18.center) to (14.center);
		\draw [thick, bend right=90, looseness=1.50] (15.center) to (19.center);
		\draw [thick] (20.center) to (21.center);
		\draw [thick] (23.center) to (22.center);
		\draw [thick, bend left, looseness=1.00] (27.center) to (15.center);
		\draw [thick, bend left, looseness=1.00] (29.center) to (28.center);
	\end{pgfonlayer}
\end{tikzpicture}}

\normalsize
\label{equ:mu-lift}
\end{equation}

We can now apply the coordination morphism of Equation \ref{equ:coord} to the compound type $N^r\ten S$ and examine how does this translate in terms of the atomic types $N^r$ and $S$. According to Equations \ref{equ:eta-lift} and \ref{equ:mu-lift} (and assuming the canonical extensions on identities) we have:

\begin{equation}
\footnotesize

\begin{tikzpicture}[ultra thick]
	\begin{pgfonlayer}{nodelayer}
		\node [style=none] (0) at (3, 1.25) {};
		\node [style=none] (1) at (3.5, 1.25) {};
		\node [draw, circle, minimum size=0.15 cm, fill=white, style=none] (2) at (4.5, 0.25) {};
		\node [style=none] (3) at (2.25, -0.5) {};
		\node [style=none] (4) at (3.5, 1.25) {};
		\node [style=none] (5) at (2.75, 0.75) {};
		\node [style=none] (6) at (0.5, 0.75) {};
		\node [style=none] (7) at (3.75, 1.25) {};
		\node [style=none] (8) at (-0.75, 1.75) {};
		\node [style=none] (9) at (6.25, -0.5) {};
		\node [style=none, fill=white, minimum size=0.15 cm, circle, draw] (10) at (2.25, 0.25) {};
		\node [style=none] (11) at (5, 0.75) {};
		\node [style=none] (12) at (7.5, 1.75) {};
		\node [style=none] (13) at (4, 0.75) {};
		\node [style=none] (14) at (0.5, -0.5) {};
		\node [style=none] (15) at (4.5, -0.5) {};
		\node [text depth=0.25 ex, text height=1.5 ex, style=none] (16) at (2.25, -1) {\footnotesize{$N^r$}};
		\node [text depth=0.25 ex, text height=1.5 ex, style=none] (17) at (-0.75, -1) {\footnotesize{$S^r$}};
		\node [style=none] (18) at (4, 1.75) {};
		\node [style=none] (19) at (6.25, 0.75) {};
		\node [text depth=0.25 ex, text height=1.5 ex, style=none] (20) at (6.25, -1) {\footnotesize{$S^l$}};
		\node [style=none] (21) at (1.75, 0.75) {};
		\node [style=none] (22) at (2.75, 0.75) {};
		\node [text depth=0.25 ex, text height=1.5 ex, style=none] (23) at (4.5, -1) {\footnotesize{$S$}};
		\node [style=none] (24) at (4, 0.75) {};
		\node [text depth=0.25 ex, text height=1.5 ex, style=none] (25) at (7.5, -1) {\footnotesize{$N$}};
		\node [text depth=0.25 ex, text height=1.5 ex, style=none] (26) at (0.5, -1) {\footnotesize{$N^{rr}$}};
		\node [style=none] (27) at (-0.75, -0.5) {};
		\node [style=none] (28) at (7.5, -0.5) {};
		\node [style=none] (29) at (2.75, 1.75) {};
		\node [style=none] (30) at (-8, 1.5) {};
		\node [style=none] (31) at (-9, -0.5) {};
		\node [style=none] (32) at (-8, 1.5) {};
		\node [draw, circle, minimum size=0.20 cm, fill=white, style=none, thick] (33) at (-9, 0.5) {};
		\node [style=none] (34) at (-10, 1.5) {};
		\node [style=none] (35) at (-12.75, 1.5) {};
		\node [style=none] (36) at (-12.75, -0.5) {};
		\node [style=none, text height=1.5 ex, text depth=0.25 ex] (37) at (-12.75, -1) {\footnotesize{$(N^r\otimes S)^r$}};
		\node [style=none] (38) at (-8, 1.5) {};
		\node [style=none] (39) at (-5.25, 1.5) {};
		\node [style=none] (40) at (-5.25, -0.5) {};
		\node [text depth=0.25 ex, text height=1.5 ex, style=none] (41) at (-9, -1) {\footnotesize{$(N^r\otimes S)$}};
		\node [style=none, text height=1.5 ex, text depth=0.25 ex] (42) at (-5.25, -1) {\footnotesize{$(N^r\otimes S)^l$}};
		\node [style=none] (43) at (-3, 1) {\footnotesize{$\mapsto$}};
	\end{pgfonlayer}
	\begin{pgfonlayer}{edgelayer}
		\draw [thick] (8.center) to (27.center);
		\draw [thick] (6.center) to (14.center);
		\draw [thick, -<] (10.center) to (3.center);
		\draw [thick, bend right=90, looseness=1.50] (21.center) to (5.center);
		\draw [thick, in=-165, out=90, looseness=1.00] (22.center) to (0.center);
		\draw [thick, in=15, out=-90, looseness=1.00] (18.center) to (7.center);
		\draw [thick, in=-15, out=90, looseness=1.00] (13.center) to (4.center);
		\draw [thick, in=165, out=-90, looseness=1.00] (29.center) to (1.center);
		\draw [thick, ->] (2.center) to (15.center);
		\draw [thick, bend right=90, looseness=1.50] (24.center) to (11.center);
		\draw [thick, bend left=90, looseness=2.00] (6.center) to (21.center);
		\draw [thick, bend left=90, looseness=2.00] (11.center) to (19.center);
		\draw [thick] (19.center) to (9.center);
		\draw [thick] (12.center) to (28.center);
		\draw [thick, bend left=90, looseness=1.25] (8.center) to (29.center);
		\draw [thick, bend left=90, looseness=1.25] (18.center) to (12.center);
		\draw [ultra thick] (33.center) to (31.center);
		\draw [ultra thick, bend right=90, looseness=1.50] (34.center) to (30.center);
		\draw [ultra thick] (35.center) to (36.center);
		\draw [ultra thick, bend left=90, looseness=2.00] (35.center) to (34.center);
		\draw [ultra thick, in=90, out=-90, looseness=1.00] (39.center) to (40.center);
		\draw [ultra thick, bend right=90, looseness=2.00] (39.center) to (38.center);
	\end{pgfonlayer}
\end{tikzpicture}}

\normalsize
\label{equ:vp-coord}
\end{equation}

Symbolically, the morphism in its expanded form becomes:

\vspace{-0.4cm}
\begin{eqnarray}
  \ol{conj}_{N^r\ten S} & = &
  (1_{(N^r\ten S)^r}\ten \mu_{N^r\ten S} \ten 1_{(N^r\ten S)^l}) \circ
  (\eta^r_{N^r\ten S} \ten \eta^l_{N^r\ten S})
   \\
  &= &
  (1_{S^r}\ten 1_{N^{rr}} \ten \mu_{N^r} \ten \mu_S \ten 1_{S^l} \ten 1_N)
  \circ
  (1_{S^r}\ten \eta^r_{N^r} \ten \sigma_{N^r,S}\ten \eta^l_S\ten 1_N) 
  \circ
  (\eta^r_S \ten \eta^r_N) 
  \nonumber
\end{eqnarray}

Putting the coordinator in context results in the following interaction:

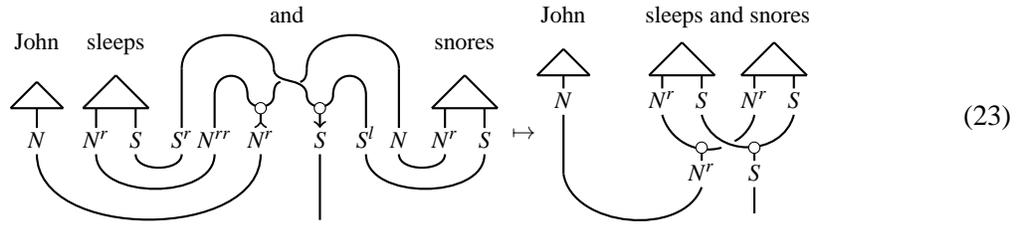
\begin{equation}
\label{equ:vp-coord1}
\footnotesize

\InputIfFileExists{./tikz/coordination1-new.tikz}{}{\input{.//tikz//coordination1-new.tikz}}

\normalsize
\end{equation}

\noindent
At the left-hand side of the diagram, we observe the following sequence of actions: 

\begin{enumerate}
  \item The subject of the coordinate structure (`John') is copied at the $N^r$ input of the coordinator; 
  \item the first branch interacts with verb `sleeps' and the second one with verb `snores'; and 
  \item the $S$ wires of the two verbs that carry the individual results are merged together with $\mu$-composition in order to return a single vector.
\end{enumerate}  

Thus the morphism of Equation \ref{equ:coord} makes effective use of both notions of copying and merging in order to represent coordination between verb phrases in a way that makes linguistic sense. Note that the simplified diagram at the right-hand side of Equation \ref{equ:vp-coord1} makes apparent a compact form for the coordinate structure ``sleeps and snores''. In fact, linear-algebraically this is nothing more than the Hadamard product between the matrices corresponding to the semantic representations of the two verb phrases. In the general case, the Hadamard product between two matrices $A=\sum_{ij}a_{ij}\ov{v_i}\ten\ov{w_j}$ and $B=\sum_{ij}b_{ij}\ov{v_i}\ten\ov{w_j}$ in $V\ten W$ is given by $A\odot B=\sum_{ij}a_{ij}b_{ij}\ov{v_i}\ten\ov{w_j}$. The element-wise merging of the two conjuncts is consistent with the way coordination is carried out over atomic types and makes possible the following convenient closed form for computing the meaning of the sentence:

\begin{equation}
  \ov{John}^{\mathsf{T}} \times (\ol{sleep}\odot\ol{snore})
\end{equation}

Although the above exposition was based on compound types of order 2, the morphism of Equation \ref{equ:coord} generalizes coherently to tensors of higher order, providing a first-class coordination object for categorical compositional distributional semantics. In the case of coordination between ditransitive verbs, for example, like in the sentence ``The bank granted Mary but denied John a loan'',  the meaning of the sentence becomes:

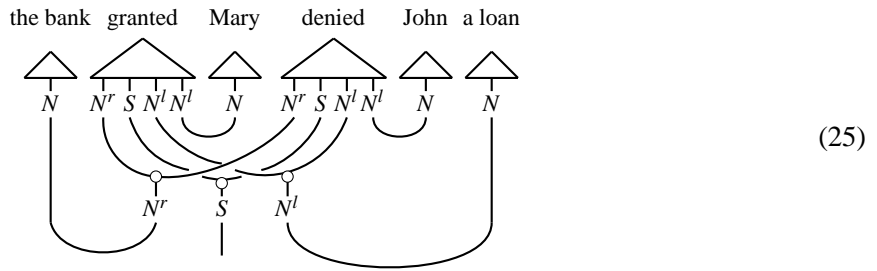
\begin{equation}
\footnotesize

\begin{tikzpicture}
	\begin{pgfonlayer}{nodelayer}
		\node [style=none] (0) at (40, -0) {};
		\node [style=none] (1) at (33.75, -2.25) {};
		\node [style=none] (2) at (32.75, 1.5) {};
		\node [style=none] (3) at (29.75, -0) {};
		\node [style=none] (4) at (29.75, -4) {};
		\node [style=none, fill=white, minimum size=0.15 cm, circle, draw] (5) at (33.75, -2.25) {};
		\node [style=none] (6) at (33.75, -2.25) {};
		\node [style=none] (7) at (33.25, 3) {};
		\node [style=none, text height=1.5 ex, text depth=0.25 ex] (8) at (31.75, 0.5) {\footnotesize{$N^r$}};
		\node [style=none] (9) at (28.75, 1.5) {};
		\node [text depth=0.25 ex, text height=1.5 ex, style=none] (10) at (29.75, 0.5) {\footnotesize{$N$}};
		\node [style=none] (11) at (33.75, -4) {};
		\node [style=none] (12) at (33.75, -3) {};
		\node [style=none] (13) at (31.75, 1) {};
		\node [style=none] (14) at (31.75, -0) {};
		\node [style=none] (15) at (31.75, 1.5) {};
		\node [style=none] (16) at (29.75, 1) {};
		\node [style=none] (17) at (30.75, 1.5) {};
		\node [text depth=0.25 ex, text height=1.5 ex, style=none] (18) at (29.75, 3.75) {the bank};
		\node [style=none] (19) at (35.25, 1.5) {};
		\node [style=none] (20) at (29.75, 1.5) {};
		\node [text depth=0.25 ex, text height=1.5 ex, style=none] (21) at (33.75, -3.5) {\footnotesize{$N^r$}};
		\node [style=none] (22) at (31.25, 1.5) {};
		\node [style=none, text height=1.5 ex, text depth=0.25 ex] (23) at (32.75, 0.5) {\footnotesize{$S$}};
		\node [style=none] (24) at (29.75, 2.5) {};
		\node [style=none] (25) at (32.75, -0) {};
		\node [style=none] (26) at (39, -0) {};
		\node [style=none, text height=1.5 ex, text depth=0.25 ex] (27) at (33.25, 3.75) {granted};
		\node [style=none] (28) at (32.75, 1) {};
		\node [style=none] (29) at (35.75, 1.5) {};
		\node [style=none] (30) at (37.75, 1.5) {};
		\node [style=none, text height=1.5 ex, text depth=0.25 ex] (31) at (36.75, 3.75) {Mary};
		\node [style=none] (32) at (36.75, 1.5) {};
		\node [style=none] (33) at (36.75, 2.5) {};
		\node [style=none] (34) at (36.75, 1) {};
		\node [style=none] (35) at (36.75, -0) {};
		\node [style=none, text height=1.5 ex, text depth=0.25 ex] (36) at (36.75, 0.5) {\footnotesize{$N$}};
		\node [style=none] (37) at (34.75, 1.5) {};
		\node [style=none] (38) at (34.75, -0) {};
		\node [text depth=0.25 ex, text height=1.5 ex, style=none] (39) at (34.75, 0.5) {\footnotesize{$N^l$}};
		\node [style=none] (40) at (33.75, 1) {};
		\node [style=none] (41) at (34.75, 1) {};
		\node [style=none] (42) at (33.75, 1.5) {};
		\node [text depth=0.25 ex, text height=1.5 ex, style=none] (43) at (33.75, 0.5) {\footnotesize{$N^l$}};
		\node [text depth=0.25 ex, text height=1.5 ex, style=none] (44) at (40, 0.5) {\footnotesize{$S$}};
		\node [text depth=0.25 ex, text height=1.5 ex, style=none] (45) at (39, 0.5) {\footnotesize{$N^r$}};
		\node [style=none] (46) at (41, 1) {};
		\node [style=none] (47) at (42.5, 1.5) {};
		\node [style=none] (48) at (41, 1.5) {};
		\node [style=none, text height=1.5 ex, text depth=0.25 ex] (49) at (42, 0.5) {\footnotesize{$N^l$}};
		\node [style=none] (50) at (40, -0) {};
		\node [style=none] (51) at (42, 1.5) {};
		\node [style=none, text height=1.5 ex, text depth=0.25 ex] (52) at (41, 0.5) {\footnotesize{$N^l$}};
		\node [style=none] (53) at (39, 1.5) {};
		\node [style=none] (54) at (41, -0) {};
		\node [style=none] (55) at (42, -0) {};
		\node [style=none] (56) at (39, -0) {};
		\node [style=none] (57) at (40.5, 3) {};
		\node [style=none] (58) at (38.5, 1.5) {};
		\node [style=none] (59) at (42, 1) {};
		\node [style=none] (60) at (39, 1) {};
		\node [style=none] (61) at (40, 1.5) {};
		\node [style=none] (62) at (40, 1) {};
		\node [style=none] (63) at (33.75, -0) {};
		\node [style=none] (64) at (41, -0) {};
		\node [text depth=0.25 ex, text height=1.5 ex, style=none] (65) at (44, 3.75) {John};
		\node [style=none] (66) at (44, 2.5) {};
		\node [style=none] (67) at (44, 0) {};
		\node [style=none] (68) at (43, 1.5) {};
		\node [style=none] (69) at (44, 1) {};
		\node [style=none] (70) at (45, 1.5) {};
		\node [text depth=0.25 ex, text height=1.5 ex, style=none] (71) at (44, 0.5) {\footnotesize{$N$}};
		\node [style=none] (72) at (44, 1.5) {};
		\node [text depth=0.25 ex, text height=1.5 ex, style=none] (73) at (46.5, 3.75) {a loan};
		\node [style=none] (74) at (46.5, 2.5) {};
		\node [style=none] (75) at (46.5, 0) {};
		\node [style=none] (76) at (45.5, 1.5) {};
		\node [style=none] (77) at (46.5, 1) {};
		\node [style=none] (78) at (47.5, 1.5) {};
		\node [text depth=0.25 ex, text height=1.5 ex, style=none] (79) at (46.5, 0.5) {\footnotesize{$N$}};
		\node [style=none] (80) at (46.5, 1.5) {};
		\node [text depth=0.25 ex, text height=1.5 ex, style=none] (81) at (40.5, 3.75) {denied};
		\node [style=none] (82) at (44, -0) {};
		\node [style=none] (83) at (42, -0) {};
		\node [style=none] (84) at (41, -0) {};
		\node [style=none] (85) at (36.25, -1.75) {};
		\node [style=none] (86) at (33.75, -0) {};
		\node [style=none] (87) at (33.75, -0) {};
		\node [style=none] (88) at (36.75, -2) {};
		\node [style=none] (89) at (32.75, -0) {};
		\node [style=none] (90) at (37.75, -2) {};
		\node [style=none] (91) at (40, -0) {};
		\node [style=none] (92) at (40, -0) {};
		\node [style=none] (93) at (35, -2) {};
		\node [style=none] (94) at (35.5, -2.25) {};
		\node [style=none] (95) at (37, -2.25) {};
		\node [draw, circle, minimum size=0.15 cm, fill=white, style=none] (96) at (36.25, -2.5) {};
		\node [style=none] (97) at (36.25, -4) {};
		\node [style=none, text height=1.5 ex, text depth=0.25 ex] (98) at (36.25, -3.5) {\footnotesize{$S$}};
		\node [style=none] (99) at (36.25, -2.5) {};
		\node [style=none] (100) at (36.25, -3) {};
		\node [style=none] (101) at (36.25, -2.5) {};
		\node [style=none] (102) at (36.25, -5.25) {};
		\node [style=none, fill=white, minimum size=0.15 cm, circle, draw] (103) at (38.75, -2.25) {};
		\node [style=none] (104) at (38.75, -4) {};
		\node [text depth=0.25 ex, text height=1.5 ex, style=none] (105) at (38.75, -3.5) {\footnotesize{$N^l$}};
		\node [style=none] (106) at (38.75, -3) {};
		\node [style=none] (107) at (46.5, -4) {};
		\node [style=none] (108) at (46.5, -0) {};
		\node [style=none] (109) at (38.75, -4) {};
	\end{pgfonlayer}
	\begin{pgfonlayer}{edgelayer}
		\draw [thick] (15.center) to (13.center);
		\draw [thick] (2.center) to (28.center);
		\draw [thick] (22.center) to (7.center);
		\draw [thick] (22.center) to (19.center);
		\draw [thick] (7.center) to (19.center);
		\draw [thick] (5.center) to (12.center);
		\draw [thick, bend right=45, looseness=1.00] (14.center) to (1.center);
		\draw [thick, bend right, looseness=0.75] (6.center) to (26.center);
		\draw [thick] (9.center) to (24.center);
		\draw [thick] (9.center) to (17.center);
		\draw [thick] (24.center) to (17.center);
		\draw [thick] (20.center) to (16.center);
		\draw [thick] (3.center) to (4.center);
		\draw [thick, bend right=75, looseness=1.00] (4.center) to (11.center);
		\draw [thick] (29.center) to (33.center);
		\draw [thick] (29.center) to (30.center);
		\draw [thick] (33.center) to (30.center);
		\draw [thick] (32.center) to (34.center);
		\draw [thick] (42.center) to (40.center);
		\draw [thick] (37.center) to (41.center);
		\draw [thick] (53.center) to (60.center);
		\draw [thick] (61.center) to (62.center);
		\draw [thick] (58.center) to (57.center);
		\draw [thick] (58.center) to (47.center);
		\draw [thick] (57.center) to (47.center);
		\draw [thick] (48.center) to (46.center);
		\draw [thick] (51.center) to (59.center);
		\draw [thick] (68.center) to (66.center);
		\draw [thick] (68.center) to (70.center);
		\draw [thick] (66.center) to (70.center);
		\draw [thick] (72.center) to (69.center);
		\draw [thick] (76.center) to (74.center);
		\draw [thick] (76.center) to (78.center);
		\draw [thick] (74.center) to (78.center);
		\draw [thick] (80.center) to (77.center);
		\draw [thick, bend right=90, looseness=1.25] (38.center) to (35.center);
		\draw [thick, bend right=90, looseness=1.25] (83.center) to (82.center);
		\draw [thick, bend left=45, looseness=1.00] (84.center) to (88.center);
		\draw [thick, bend left, looseness=0.75] (85.center) to (87.center);
		\draw [thick, bend right, looseness=1.00] (89.center) to (93.center);
		\draw [thick, bend right, looseness=0.75] (90.center) to (92.center);
		\draw [thick, bend right=15, looseness=1.00] (94.center) to (95.center);
		\draw [thick] (96.center) to (100.center);
		\draw [thick] (97.center) to (102.center);
		\draw [thick] (103.center) to (106.center);
		\draw [thick] (108.center) to (107.center);
		\draw [thick, bend left=90, looseness=0.75] (107.center) to (109.center);
	\end{pgfonlayer}
\end{tikzpicture}}

\normalsize
\end{equation}

Note that the above derivation accounts for copying {\em both} the subject and the direct object, as required by distributivity. This is translated to the following formula:

\begin{equation}
  \ov{bank}^{\mathsf{T}} \times \left[ (\ol{grant} \times \ov{Mary}) \odot (\ol{deny} \times \ov{John}) \right] \times \ov{loan}
\end{equation}

\noindent
where $\ol{grant}$ and $\ol{deny}$ refer to tensors of order 4, and the $\odot$ symbol denotes element-wise multiplication between tensors of order 3.

\subsection{Non-standard forms of coordination}
\label{sec:stripping}

In this section we briefly examine the use of Frobenius operators in non-standard forms of coordination. Consider for example the following sentence:

\begin{exe}
  \ex\label{ex:stripping} John likes Poe, and$_{[\text{John likes}]}$ Lovecraft as well
\end{exe}

Example \ref{ex:stripping} exhibits a form of ellipsis related to coordination known as {\em stripping}; we expect that the meaning of this sentence will be the same with that of the standard form ``John likes Poe and Lovercraft'', which is computed below according to Equation \ref{equ:coord}:

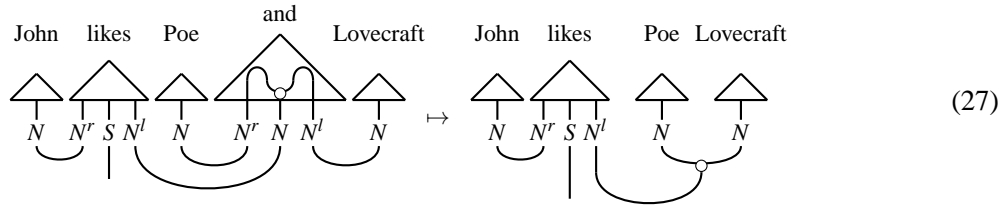
\begin{equation}
\footnotesize

\begin{tikzpicture}
	\begin{pgfonlayer}{nodelayer}
		\node [style=none, text height=1.5 ex, text depth=0.25 ex] (0) at (-3, 3.25) {and};
		\node [style=none, text height=1.5 ex, text depth=0.25 ex] (1) at (-12.25, 2.5) {John};
		\node [style=none, text height=1.5 ex, text depth=0.25 ex] (2) at (-9.5, 2.5) {likes};
		\node [style=none, text height=1.5 ex, text depth=0.25 ex] (3) at (0.75, 2.5) {Lovecraft};
		\node [style=none] (4) at (-3, 2.5) {};
		\node [style=none] (5) at (-9.5, 1.5) {};
		\node [style=none] (6) at (-12.25, 1) {};
		\node [style=none] (7) at (-6.75, 1) {};
		\node [style=none] (8) at (-4.25, 0.75) {};
		\node [style=none] (9) at (-3.5, 0.75) {};
		\node [style=none] (10) at (-2.5, 0.75) {};
		\node [style=none] (11) at (-1.75, 0.75) {};
		\node [draw, circle, minimum size=0.15 cm, fill=white, style=none] (12) at (-3, 0.25) {};
		\node [style=none] (13) at (-13.25, -0) {};
		\node [style=none] (14) at (-12.25, -0) {};
		\node [style=none] (15) at (-11.25, -0) {};
		\node [style=none] (16) at (-11, -0) {};
		\node [style=none] (17) at (-10.5, -0) {};
		\node [style=none] (18) at (-9.5, -0) {};
		\node [style=none] (19) at (-8.5, -0) {};
		\node [style=none] (20) at (-8, -0) {};
		\node [style=none] (21) at (-7.75, -0) {};
		\node [style=none] (22) at (-6.75, -0) {};
		\node [style=none] (23) at (-5.75, -0) {};
		\node [style=none] (24) at (-5.5, -0) {};
		\node [style=none] (25) at (-4.25, -0) {};
		\node [style=none] (26) at (-1.75, -0) {};
		\node [style=none] (27) at (-0.5, -0) {};
		\node [style=none] (28) at (-0.25, -0) {};
		\node [style=none] (29) at (-12.25, -0.75) {};
		\node [style=none] (30) at (-10.5, -0.75) {};
		\node [style=none] (31) at (-9.5, -0.75) {};
		\node [style=none] (32) at (-8.5, -0.75) {};
		\node [style=none] (33) at (-6.75, -0.75) {};
		\node [style=none] (34) at (-4.25, -0.75) {};
		\node [style=none] (35) at (-4.25, -0.75) {};
		\node [style=none] (36) at (-3, -0.75) {};
		\node [style=none] (37) at (-3, -0.75) {};
		\node [style=none] (38) at (-1.75, -0.75) {};
		\node [style=none] (39) at (-1.75, -0.75) {};
		\node [style=none, text height=1.5 ex, text depth=0.25 ex] (40) at (-12.25, -1.25) {\footnotesize{$N$}};
		\node [style=none, text height=1.5 ex, text depth=0.25 ex] (41) at (-10.5, -1.25) {\footnotesize{$N^r$}};
		\node [style=none, text height=1.5 ex, text depth=0.25 ex] (42) at (-9.5, -1.25) {\footnotesize{$S$}};
		\node [style=none, text height=1.5 ex, text depth=0.25 ex] (43) at (-8.5, -1.25) {\footnotesize{$N^l$}};
		\node [style=none, text height=1.5 ex, text depth=0.25 ex] (44) at (-6.75, -1.25) {\footnotesize{$N$}};
		\node [style=none, text height=1.5 ex, text depth=0.25 ex] (45) at (-4.25, -1.25) {\footnotesize{$N^r$}};
		\node [style=none, text height=1.5 ex, text depth=0.25 ex] (46) at (-3, -1.25) {\footnotesize{$N$}};
		\node [style=none, text height=1.5 ex, text depth=0.25 ex] (47) at (-1.75, -1.25) {\footnotesize{$N^l$}};
		\node [style=none] (48) at (-12.25, -1.75) {};
		\node [style=none] (49) at (-10.5, -1.75) {};
		\node [style=none] (50) at (-8.5, -1.75) {};
		\node [style=none] (51) at (-6.75, -1.75) {};
		\node [style=none] (52) at (-4.25, -1.75) {};
		\node [style=none] (53) at (-9.5, -1.75) {};
		\node [style=none] (54) at (-1.75, -1.75) {};
		\node [style=none] (55) at (-9.5, -3) {};
		\node [text depth=0.25 ex, text height=1.5 ex, style=none] (56) at (-6.75, 2.5) {Poe};
		\node [style=none] (57) at (0.75, -1.75) {};
		\node [style=none] (58) at (1.75, -0) {};
		\node [style=none] (59) at (0.75, -0.75) {};
		\node [text depth=0.25 ex, text height=1.5 ex, style=none] (60) at (0.75, -1.25) {\footnotesize{$N$}};
		\node [style=none] (61) at (-0.25, -0) {};
		\node [style=none] (62) at (0.75, 1) {};
		\node [style=none] (63) at (0.75, -0) {};
		\node [style=none] (64) at (-1.75, -1.75) {};
		\node [style=none] (65) at (0.75, -1.75) {};
		\node [style=none] (66) at (-3, -1.75) {};
		\node [style=none] (67) at (8, -0.75) {};
		\node [style=none] (68) at (5.25, -0) {};
		\node [style=none] (69) at (11.5, -0.75) {};
		\node [style=none] (70) at (7, -1.75) {};
		\node [style=none] (71) at (9, -0.75) {};
		\node [style=none] (72) at (13.5, -0) {};
		\node [text depth=0.25 ex, text height=1.5 ex, style=none] (73) at (9, -1.25) {\footnotesize{$N^l$}};
		\node [style=none] (74) at (5.25, -1.75) {};
		\node [style=none] (75) at (9, -2.75) {};
		\node [style=none] (76) at (9, -0) {};
		\node [text depth=0.25 ex, text height=1.5 ex, style=none] (77) at (14.5, 2.5) {Lovecraft};
		\node [style=none] (78) at (7, -0) {};
		\node [style=none] (79) at (13.5, -0) {};
		\node [style=none] (80) at (4.25, -0) {};
		\node [style=none] (81) at (11.5, -0) {};
		\node [style=none] (82) at (14.5, -1.75) {};
		\node [style=none] (83) at (9.5, -0) {};
		\node [style=none] (84) at (14.5, 1) {};
		\node [style=none] (85) at (8, 1.5) {};
		\node [style=none] (86) at (14.5, -0) {};
		\node [style=none] (87) at (8, -0) {};
		\node [style=none] (88) at (6.5, -0) {};
		\node [style=none] (89) at (9, -1.75) {};
		\node [text depth=0.25 ex, text height=1.5 ex, style=none] (90) at (11.5, -1.25) {\footnotesize{$N$}};
		\node [style=none] (91) at (11.5, -1.75) {};
		\node [style=none] (92) at (6.25, -0) {};
		\node [style=none, text height=1.5 ex, text depth=0.25 ex] (93) at (14.5, -1.25) {\footnotesize{$N$}};
		\node [style=none] (94) at (12.5, -0) {};
		\node [text depth=0.25 ex, text height=1.5 ex, style=none] (95) at (8, -1.25) {\footnotesize{$S$}};
		\node [style=none] (96) at (14.5, -0.75) {};
		\node [style=none, text height=1.5 ex, text depth=0.25 ex] (97) at (11.5, 2.5) {Poe};
		\node [style=none] (98) at (15.5, -0) {};
		\node [style=none] (99) at (11.5, 1) {};
		\node [text depth=0.25 ex, text height=1.5 ex, style=none] (100) at (5.25, 2.5) {John};
		\node [style=none] (101) at (9, -2.75) {};
		\node [style=none] (102) at (10.5, -0) {};
		\node [style=none] (103) at (5.25, -0.75) {};
		\node [style=none] (104) at (13, -2.75) {};
		\node [style=none] (105) at (5.25, 1) {};
		\node [text depth=0.25 ex, text height=1.5 ex, style=none] (106) at (8, 2.5) {likes};
		\node [text depth=0.25 ex, text height=1.5 ex, style=none] (107) at (7, -1.25) {\footnotesize{$N^r$}};
		\node [style=none] (108) at (7, -0.75) {};
		\node [text depth=0.25 ex, text height=1.5 ex, style=none] (109) at (5.25, -1.25) {\footnotesize{$N$}};
		\node [style=none] (110) at (13, -2.75) {};
		\node [style=none, fill=white, minimum size=0.15 cm, circle, draw] (111) at (13, -2.5) {};
		\node [style=none] (112) at (8, -3.75) {};
		\node [style=none] (113) at (8, -1.75) {};
		\node [style=none] (114) at (8, -3.25) {};
		\node [style=none] (115) at (3, -0.75) {$\mapsto$};
	\end{pgfonlayer}
	\begin{pgfonlayer}{edgelayer}
		\draw [thick, bend left=90, looseness=2.25] (8.center) to (9.center);
		\draw [thick] (22.center) to (33.center);
		\draw [thick] (24.center) to (4.center);
		\draw [thick] (19.center) to (32.center);
		\draw [thick, bend right=90, looseness=1.50] (9.center) to (10.center);
		\draw [thick] (18.center) to (31.center);
		\draw [thick] (14.center) to (29.center);
		\draw [thick] (13.center) to (6.center);
		\draw [thick] (16.center) to (5.center);
		\draw [thick] (12.center) to (36.center);
		\draw [thick] (11.center) to (26.center);
		\draw [thick, bend right=90, looseness=1.00] (51.center) to (52.center);
		\draw [thick] (7.center) to (23.center);
		\draw [thick] (5.center) to (20.center);
		\draw [thick] (25.center) to (35.center);
		\draw [thick] (24.center) to (27.center);
		\draw [thick, bend right=90, looseness=1.00] (48.center) to (49.center);
		\draw [thick] (17.center) to (30.center);
		\draw [thick] (21.center) to (7.center);
		\draw [thick] (21.center) to (23.center);
		\draw [thick] (13.center) to (15.center);
		\draw [thick] (53.center) to (55.center);
		\draw [thick] (26.center) to (38.center);
		\draw [thick] (6.center) to (15.center);
		\draw [thick] (16.center) to (20.center);
		\draw [thick] (8.center) to (25.center);
		\draw [thick] (4.center) to (27.center);
		\draw [thick, bend left=90, looseness=2.25] (10.center) to (11.center);
		\draw [thick] (63.center) to (59.center);
		\draw [thick] (61.center) to (62.center);
		\draw [thick] (61.center) to (58.center);
		\draw [thick] (62.center) to (58.center);
		\draw [thick, bend right=90, looseness=1.00] (64.center) to (65.center);
		\draw [thick, bend right=90, looseness=1.00] (50.center) to (66.center);
		\draw [thick] (81.center) to (69.center);
		\draw [thick] (76.center) to (71.center);
		\draw [thick] (87.center) to (67.center);
		\draw [thick] (68.center) to (103.center);
		\draw [thick] (80.center) to (105.center);
		\draw [thick] (88.center) to (85.center);
		\draw [thick, bend right=90, looseness=0.75] (91.center) to (82.center);
		\draw [thick] (99.center) to (94.center);
		\draw [thick] (85.center) to (83.center);
		\draw [thick, bend right=90, looseness=1.00] (74.center) to (70.center);
		\draw [thick] (78.center) to (108.center);
		\draw [thick] (102.center) to (99.center);
		\draw [thick] (102.center) to (94.center);
		\draw [thick] (80.center) to (92.center);
		\draw [thick] (89.center) to (75.center);
		\draw [thick] (105.center) to (92.center);
		\draw [thick] (88.center) to (83.center);
		\draw [thick] (86.center) to (96.center);
		\draw [thick] (79.center) to (84.center);
		\draw [thick] (79.center) to (98.center);
		\draw [thick] (84.center) to (98.center);
		\draw [thick, bend right=90, looseness=1.00] (101.center) to (104.center);
		\draw [thick] (111.center) to (110.center);
		\draw [thick] (113.center) to (112.center);
	\end{pgfonlayer}
\end{tikzpicture}}

\normalsize
\end{equation}

As a special form of coordination, stripping can be addressed by exploiting a special-purpose tensor; we achieve the desired result by defining the internal structure of the coordinator as shown in the following diagram:

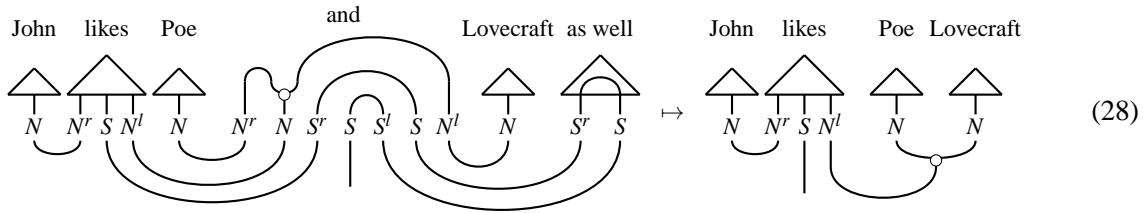
\vspace{-0.5cm}
\begin{equation}
\footnotesize

\InputIfFileExists{./tikz/strip3.tikz}{}{\input{.//tikz//strip3.tikz}}

\normalsize
\end{equation}

Note that the actual coordination step is still carried out by the morphism of Equation \ref{equ:coord}; however, additional wiring is necessary in order to allow the unhindered flow of information through the left-hand and right-hand parts of the sentence.

\section{Conclusion and future work}

This paper contributes to the ongoing research on categorical compositional distributional semantics by providing an account of coordination. While the presented ideas cover the most common cases of coordination, language is more complex than that; for example, coordination between text constituents of different syntactic categories (``John works evenings and on weekeends'') constitutes an interesting future direction of research. A proper distinction between conjunction and disjunction cases (which is not pursued in this paper) remains an open problem, as it implies the presence of an underlying logic; to this end, applications of quantum logic seem to provide a promising direction, as the work of Widdows \cite{widdows2003orthogonal} and Van Rijsbergen \cite{rijsbergen} shows. Finally, the Frobenius machinery and especially the copying operator provides a means for addressing other forms of ellipsis in language, such as verb phrase ellipsis or gapping, that we plan to explore in the future. 

\vspace{-0.2cm}
\section*{Acknowledgements}
\vspace{-0.2cm}

The author would like to thank the reviewers for their valuable comments, as well as Bob Coecke, Mehrnoosh Sadrzadeh and Matt Purver for useful suggestions and discussions. Support by the AFOSR grant ``Algorithmic and Logical Aspects when Computing Meaning'' is gratefully acknowledged.

\bibliographystyle{eptcs}
\bibliography{refs}
\end{document}